\documentclass[10pt,twocolumn,letterpaper]{article}
\usepackage{3dv}
\usepackage{float}
\usepackage{amsmath}
\usepackage{amssymb}
\usepackage{graphicx}
\usepackage{tabularx}
\usepackage{times}

\DeclareMathOperator{\tr}{tr}

\PassOptionsToPackage{hyphens}{url}\usepackage[pagebackref=true,breaklinks=true,letterpaper=true,colorlinks,bookmarks=false]{hyperref}

\graphicspath{{figures/}{images/}}

\threedvfinalcopy 

\def\threedvPaperID{123}  

\ifthreedvfinal\fi
\begin{document}

\title{A Progressive Conditional Generative Adversarial Network\\ 
for Generating Dense and Colored 3D Point Clouds}

\author{Mohammad Samiul Arshad and William J. Beksi\\
Department of Computer Science and Engineering\\ 
University of Texas at Arlington\\ 
Arlington, TX, USA\\
{\tt\small mohammadsamiul.arshad@mavs.uta.edu}, {\tt\small william.beksi@uta.edu}
}

\maketitle

\begin{abstract}
In this paper, we introduce a novel conditional generative adversarial network 
that creates dense 3D point clouds, with color, for assorted classes of objects 
in an unsupervised manner. To overcome the difficulty of capturing intricate 
details at high resolutions, we propose a point transformer that progressively 
grows the network through the use of graph convolutions. The network is composed 
of a leaf output layer and an initial set of branches. Every training iteration 
evolves a point vector into a point cloud of increasing resolution. After a 
fixed number of iterations, the number of branches is increased by replicating 
the last branch. Experimental results show that our network is capable of 
learning and mimicking a 3D data distribution, and produces colored point clouds 
with fine details at multiple resolutions. 
\end{abstract}

\vspace{-1mm}
\section{Introduction} 
\vspace{-1mm}
\label{sec:introduction}
In recent years, research on processing 3D point clouds has gained momentum due 
to an increasing number of relevant applications. From robot navigation 
\cite{biswas2012depth, luo20163d} to autonomous vehicles \cite{pfrunder2017real,
whitty2010autonomous,zhu20123d}, augmented reality \cite{wang2018point,
pfrunder2017real} to health care \cite{liu20183d,bostelman2006applications,
pohlmann2016evaluation}, the challenges of working with 3D datasets are being 
realized. Among miscellaneous data modalities, raw point clouds are becoming 
popular as a compact homogeneous representation that has the ability to capture 
intricate details of the environment. Intuitively, a 3D point cloud can be 
thought of as an unordered set of irregular points collected from the surface of 
an object. Each point consists of a Cartesian coordinate, along with other 
additional information such as a surface normal estimate and RGB color value. 
Although 3D point clouds are the product of range sensing devices (e.g.,
structured light, time-of-flight, light detection and ranging, etc.), the 
application of conventional machine learning techniques on the direct sensor 
output is nontrivial. In particular, deep learning methods fall short in the 
processing of 3D point clouds due to the irregular and permutation invariant 
nature of the data. 

\begin{figure}[t]
\begin{center}
  \includegraphics[width=0.8\linewidth]{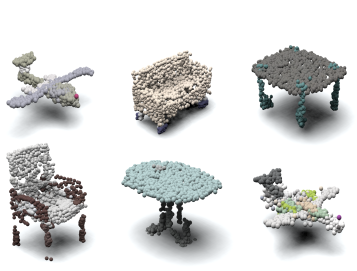}
\end{center}
\caption{Examples of 3D point clouds synthesized by our progressive conditional 
generative adversarial network (PCGAN) for an assortment of classes. PCGAN 
generates both geometry and color for point clouds, without supervision, 
through a coarse to fine training process. Best viewed in color.}
\label{fig:pcgan_results_1}
\vspace{-2mm}
\end{figure}

Generating synthetic 3D point cloud data is an open area of research with the 
intention of facilitating the learning of non-Euclidean point representations. 
In three dimensions, synthetic data may take the form of meshes, voxels, or raw 
point clouds in order to learn a representation that aids the solution of 
computer vision tasks such as classification \cite{qi2017pointnet,
wang2019dynamic,maturana2015voxnet,zhou2018voxelnet,cao20173d}, segmentation 
\cite{qi2017pointnet,wang2019dynamic,qi2017pointnet++,yi2017syncspeccnn,
hou20193d,yi2019gspn,sharma2019learning,wu2019pointconv}, and reconstruction 
\cite{wu2018learning,wang2019forknet,wen2020point,sarmad2019rl,yuan2018pcn,
tchapmi2019topnet}. Currently, researchers make use of point clouds sampled 
from the mesh of manually designed objects as synthetic data for training deep 
learning models \cite{qi2017pointnet,wang2019dynamic,shu20193d,cao2019point}. 
However, the geometry and texture of these point clouds is bounded by the 
resolution of the modeled objects. Moreover, due to the complexity of the design 
process, the number of composed objects can fail to satisfy the enormous data 
needs of deep learning research. Automatically synthesizing 3D point clouds can 
solve this problem by providing a source of potentially infinite amounts of 
diverse data. 

Although color and geometry are among the defining features of 3D objects in 
the physical world, current point cloud generation methods either create the 
geometry \cite{shu20193d,yang2019pointflow,valsesia2018learning, 
achlioptas2018learning,hertz2020pointgmm,ramasinghe2019spectral} or the color 
\cite{cao2019point} of the point cloud, but not both. We believe that 3D point 
cloud generators should have the ability to synthesize dense point clouds with 
complex details that mimic real-world objects in both {\em geometry} and 
{\em color}. To this end, we propose a progressive conditional generative 
adversarial network (PCGAN) that faithfully generates dense point clouds with 
color using tree structured graph convolutions. To the best of our knowledge, 
our work is the first attempt to generate both geometry and color for point 
clouds in an unsupervised fashion with progressive, i.e., coarse to fine 
training. Figure \ref{fig:pcgan_results_1} shows examples of 3D point clouds 
generated by our network. 

Generating high-resolution data is a difficult and computationally expensive 
task as the number of features and the level of details increases with 
resolution. In traditional generative methods \cite{goodfellow2014generative, 
arjovsky2017wasserstein,gulrajani2017improved}, the generator tries to optimize 
both global structure and local features simultaneously which can overwhelm an 
unsupervised network. To reduce the learning complexity, a progressively growing 
generative adversarial network \cite{karras2017progressive} may be used to learn 
features in a coarse to fine manner. Progressive growing has also been shown to 
improve training time \cite{karras2017progressive} since the generator first 
optimizes low-resolution global structures which helps the optimization of local 
details at higher resolutions. In terms of 3D point clouds, boosting the 
resolution makes the dataset denser by the expansion of the number of points. 
Consequently, this adds more complexity to the generation procedure and 
amplifies the computational cost. This directly proportional relationship 
between point cloud resolution and complexity/computational cost has inspired us
to use progressive growing in PCGAN for dense point cloud generation. Our 
network is end-to-end trainable and it learns the distribution of the data as 
well as the mapping from label to data to generate samples of numerous classes 
with a single network.

\subsection{Contributions} 
\label{subsec:contributions}
The key contributions of our work are threefold. 
\begin{itemize}
  \setlength{\itemsep}{-\parsep}\setlength{\topsep}{-\parsep}                                                                                                 
  \item We introduce a progressive generative network that creates both geometry 
  and color for 3D point clouds in the absence of supervision.
  \item We include both a qualitative and quantitative analysis of the objects 
  synthesized by our network.
  \item We present the Fr\`echet dynamic distance metric to evaluate colored 
  dense point clouds. 
\end{itemize}
To allow other researchers to use our software, reproduce the results, and 
improve on them, we have released PCGAN under an open-source license. The source 
code and detailed installation instructions are available online 
\cite{pcgan2020}.

The remainder of this paper is organized as follows. We give a summary of 
related research in Section \ref{sec:related_work}. In Section 
\ref{sec:problem_statement}, we define the problem mathematically and provide 
the essential background information in Section \ref{sec:background}. The 
details of our model are provided in Section \ref{sec:model}, and we present the
experimental setup and results in Section \ref{sec:experiments}. A conclusion 
of this work is given in Section \ref{sec:conclusion}. 

\vspace{-1mm}
\section{Related Work} 
\vspace{-1mm}
\label{sec:related_work}
In this section, we summarize pertinent work on the generation of 3D point 
clouds. Interested readers are encouraged to read \cite{chaudhuri2019learning,
ahmed2018survey,ioannidou2017deep} for a comprehensive survey of deep learning
research on 3D point cloud datasets.

\textbf{3D Point Cloud Generation.} The first generative model capable of 
producing raw point clouds comes from the work of Achlioptas \etal 
\cite{achlioptas2018learning}. Using PointNet \cite{qi2017pointnet} as the 
backbone, Achlioptas \etal introduced an autoencoder and two variants of a
generative adversarial network to generate point clouds. Prior to 
\cite{achlioptas2018learning}, Qi \etal introduced PointNet 
\cite{qi2017pointnet}, the first neural network to operate directly on point 
cloud data. Eckart \etal \cite{eckart2016accelerated} used hierarchical 
Gaussian mixture models (hGMMs) to process point clouds, Zaheer \etal 
\cite{zaheer2017deep} utilized deep networks to analyze point clouds as sets, 
and Li \etal \cite{li2018so} made use of self-organizing maps and hierarchical 
feature extraction to discriminate point clouds.

Following \cite{achlioptas2018learning}, Li \etal \cite{li2018point} used two 
generative networks to learn a latent distribution and generated points based 
on learned features. Yang \etal \cite{yang2018foldingnet} improved upon 
\cite{achlioptas2018learning} by incorporating graph-based enhancement on top 
of PointNet and 2D grid deformation. Valsesia \etal \cite{valsesia2018learning}
used graph convolutions and exploited pairwise distances between features to 
construct a generator. Similar to \cite{li2018point}, Yang \etal 
\cite{yang2019pointflow} generated point clouds by learning two hierarchical
distributions and through the use of continuous normalizing flow. Mo \etal 
\cite{mo2019structurenet} mapped part hierarchies of an object shape as a tree 
and implemented an encoder-decoder network to generate new shapes via 
interpolation. Gadelha \etal \cite{gadelha2018multiresolution} presented a tree 
network for 3D shape understanding and generation tasks by operating on 
1D-ordered point lists obtained from a k-d tree space partitioning.

Ramasinghe \etal \cite{ramasinghe2019spectral} generated high-resolution point 
clouds by operating on the spatial domain, and Hertz \etal 
\cite{hertz2020pointgmm} used hGMMs to generate shapes in different 
resolutions. However, both \cite{ramasinghe2019spectral} and 
\cite{hertz2020pointgmm} fail to generate fine shape details. Xie \etal 
\cite{xie2020generative} proposed an energy-based generative PointNet 
\cite{qi2017pointnet}, and Sun \etal \cite{sun2020pointgrow} implemented 
auto-regressive learning with self-attention and context awareness for the 
generation and completion of point clouds. Cao \etal used adversarial training 
to generate color for point cloud geometry in \cite{cao2019point}. In follow up 
work, they used a style transfer approach to transform the geometry and color 
of a candidate point cloud according to a target point cloud or image 
\cite{cao2020psnet}. 

Compared to the focus of the aforementioned works on solely generating the 
geometry or color of point clouds, our model can produce both color and geometry 
in an unsupervised manner while maintaining exceptional details at high 
resolutions. The generator of our model is inspired by the tree structured graph 
convolution generator of Shu \etal \cite{shu20193d}. However, the generator 
proposed by Shu \etal does not generate point clouds in color nor does it 
incorporate the advantages of progressive training. The multiclass generation 
model of Shu \etal is class agnostic thus making the generation process of a 
specific class of objects uncontrollable. Conversely, we incorporate conditional 
generation to control the creation process and use progressive training to build 
high-resolution point clouds with color. Although, Tchapmi \etal 
\cite{tchapmi2019topnet} also introduced a tree graph-based decoder, the focal 
point of their work was the completion of point cloud geometry with supervision. 

\textbf{Graph Convolutions.} Point clouds can be portrayed as graphs where 
points represent nodes and the co-relation among neighboring points represent 
edges. Applying the notion of graph convolution operations to process point 
clouds is a relatively new area of research. In \cite{qi2017pointnet++}, Qi 
\etal proposed the hierarchical application of PointNet to learn point cloud 
features as a graph embedding. However, their work did not incorporate the 
co-relation of neighboring points and consequently disregards local features. To
account for local features, Atzmon \etal \cite{atzmon2018point} used a Gaussian 
kernel applied to the pairwise distances between points. Wang \etal 
\cite{wang2019dynamic} introduced DGCNN which uses aggregated point features 
and pairwise distances among $k$ points to dynamically construct a graph. The 
discriminator of our model was motivated by DGCNN. Nevertheless, DGCNN was 
designed for the classification of point cloud geometry while our discriminator 
aims to act as a critic of point cloud geometry and color to distinguish 
between real and synthetic data given a class label.

\textbf{Progressive Training.} Progressive training has been shown to improve 
the quality of 2D image generation \cite{karras2017progressive,karras2019style, 
karras2020analyzing}. Our work is the first attempt to use progressive 
training in an unsupervised 3D generative network. In previous research, 
Valsesia \etal \cite{valsesia2018learning} used upsampling layers based on $k$ 
neighbors of the adjacency graph to increase the feature size between each 
graph convolution. However, their proposed architecture learns without 
progressive growing and suffers the same computational complexity of regular 
generative adversarial networks. The work of Yifan \etal \cite{yuan2018pcn} is 
the only example of coarse-to-fine generation in 3D. Yet, Yifan \etal used a 
supervised patch-based approach to upsample point clouds where the overall 
structure of the data is provided as a prior. In contrast, our network 
progressively learns the global shape of the data through the underlying 
distribution of the point cloud geometry and color of a class with no 
supervision or priors given.

\vspace{-1mm}
\section{Problem Statement} 
\vspace{-1mm}
\label{sec:problem_statement}
Consider a set of classes, $C = \{c_1,\ldots,c_n\}$, each representing mixed 
objects as 3D colored point clouds, $x \in \mathbb{R}^{N \times 6}$, where $N$ 
is the number of points. Given a class $c \in C$, we seek to learn the 
underlying features that constitute $c$ and generate a realistic point cloud 
$\hat{x} \in \mathbb{R}^{N \times 6}$.

\vspace{-1mm}
\section{Background} 
\vspace{-1mm}
\label{sec:background}
In the following subsections we lay out the necessary background knowledge upon
which our work is grounded.

\subsection{Wasserstein or Kantorovich-Rubinstein Distance}
\label{subsec:wasserstein_distance}
Given two distributions $P_r$ and $P_g$ in a metric space $\mathcal{M}$, the 
Wasserstein or Kantorovich-Rubinstein distance calculates the minimal cost to 
transform $P_r$ into $P_g$ or vice-versa \cite{villani2008optimal}. A distance 
of order $p$ can be expressed as 
\begin{equation*}
  W_p(P_r,P_g) = \underset{\gamma \in \Pi(P_r,P_g)}{\inf}\mathbb{E}_{(x,y)\sim\gamma}\big[\|x-y\|\big],
\end{equation*}
where $\Pi(P_r,P_g)$ is the set of all joint distributions $\gamma(x,y)$ with 
marginals $P_r$ and $P_g$.

\subsection{Generative Adversarial Networks}
\label{subsec:generative_adversarial_network}
Introduced by Goodfellow \etal \cite{goodfellow2014generative}, a generative 
adversarial network (GAN) is a special type of neural network that focuses on 
learning the underlying distribution, $P_r$, of a dataset to generate new 
samples. A GAN consists of a generator $G$ and a discriminator $D$ that compete 
against each other in a minimax game. The generator tries to manipulate a random 
vector $z$, drawn from a fixed distribution $P_g$, into synthetic samples 
$\hat{x}$ that are indistinguishable from the real data $x$. The discriminator 
seeks to differentiate between $\hat{x}$ and $x$. More formally, the objective 
of a GAN can be written as 
\begin{align*}
  \underset{G}\min\,\underset{D}\max(G, D) &= \mathbb{E}_{x \sim P_r}[\log D(x)]\\
  &+ \mathbb{E}_{z \sim P_{g}}[\log(1-D(G(z)))].
\end{align*}

\subsection{Wasserstein GAN with Gradient Penalty}
\label{subsec:wasserstein_gan-gp}
To optimize the convergence of a GAN, Arjovsky \etal 
\cite{arjovsky2017wasserstein} introduced the Wasserstein GAN (WGAN) whose goal 
is to minimize the distance between the real data distribution $P_r$ and the 
generated data distribution $P_g$ using the Wasserstein metric. To make the goal 
of inter-distribution distance minimization tractable, Arjovsky \etal used the 
dual form of the Wasserstein distance, i.e., the Wasserstein-1 
\cite{villani2008optimal} with the GAN objective  
\begin{equation*}
  \underset{G}\min\,\underset{D\in\mathcal{D}}\max(G, D) = \mathbb{E}_{x \sim 
  P_r}[D(x)] - \mathbb{E}_{z \sim P_{g}}[D(G(z))],
\end{equation*}
where $\mathcal{D}$ is the set of 1-Lipschitz functions. To ensure continuity 
in space, the discriminator of the WGAN must be 1-Lipschitz which is achieved 
through weight clipping. However, since weight clipping penalizes the norm of 
the gradient the stability of network can be compromised. 

Gulrajani \etal \cite{gulrajani2017improved} improved the WGAN by using a 
gradient penalty (WGAN-GP) instead of weight clipping. To enforce the
1-Lipschitz condition, the WGAN-GP constrains the norm of the gradient to be at
most 1. This is achieved by a penalty term applied to the gradient of the 
discriminator, 
\begin{align*}
  \underset{G}\min\,\underset{D\in\mathcal{D}}\max(G, D) &= \mathbb{E}_{x \sim 
  P_r}[D(x)] - \mathbb{E}_{z \sim P_{g}}[D(G(z))]\\ 
  &+ \lambda\,\mathbb{E}_{\tilde{x}\sim 
  P_{\tilde{x}}}\big[(\|\nabla_{\tilde{x}}D(\tilde{x})\|_2 - 1)^2\big],
\end{align*}
where $\tilde{x}\sim P_{\tilde{x}}$ are the points uniformly sampled along the 
straight line between pairs of points from the real data distribution $P_r$ and 
generated data distribution $P_g$.

\vspace{-1mm}
\section{Model Architecture} 
\begin{figure*}
\begin{center}
  \includegraphics[width=0.8\linewidth]{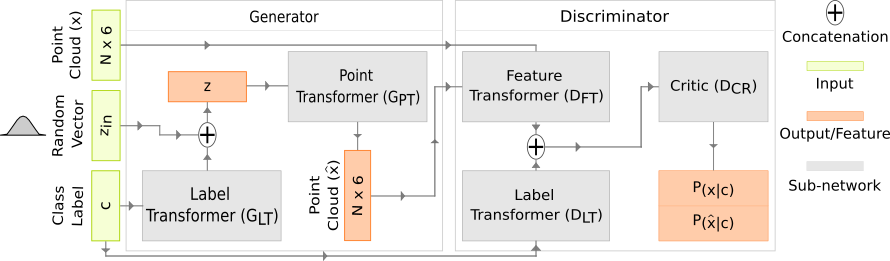}
\end{center}
\caption{Given a random vector $z_{in}$ and class label $c$, the generator 
produces a colored point cloud $\hat{x}$. The real point cloud $x$, class label 
$c$, and the generated point cloud $\hat{x}$ are given to the discriminator 
which then tries to predict the probability of the data being real or synthetic. 
Best viewed in color.}
\label{fig:pcgan_model}
\end{figure*}

\vspace{-1mm}
\label{sec:model}
Our model has the following two main components: a generator $G$ and a 
discriminator $D$. Unless stated otherwise, we refer to a point cloud $x$ as a
6-dimensional matrix with $N$ points, i.e., $x \in \mathbb{R}^{N\times 6}$ where 
each point represents a Cartesian coordinate and RGB color. For multiclass 
generation, both the generator and the discriminator are conditioned on a class 
label, $c \in C$, which is randomly chosen from a set of $n$ classes 
$C = \{c_1,\ldots,c_n\}$. To optimize the generator and the discriminator we 
make use of the WGAN-GP techniques discussed in Section 
\ref{subsec:wasserstein_gan-gp}. Figure \ref{fig:pcgan_model} shows the overall
architecture of PCGAN. 

\subsection{Generator} 
\label{subsec:generator}
The generator $G$ takes as input a random vector, $z_{in} \in \mathcal{N}(0,1)$,
along with a class label $c \in C$ represented as one hot vector. The class 
label controls the generation of colored point clouds of a desired class. The 
generator is comprised of two sub-networks: a label transformer $G_{LT}$ and a 
point transformer $G_{PT}$.

The class label goes through $G_{LT}$, a shallow two-layer perceptron, and a 
class vector $g_c \in \mathbb{R}^{64}$ is constructed. The input vector $z_{in}$ 
is then concatenated with $g_c$ to produce a point vector $z$ which is given to 
$G_{PT}$ as input, i.e., 
\begin{equation*}
  z = (z_{in}, g_c).
\end{equation*}
Following \cite{shu20193d}, the point transformer incorporates tree structured 
graph convolutions (TreeGCN). As the name suggests, information in TreeGCN is 
passed from the root node to a leaf node instead of all neighbors, i.e., the
$i$-th node at layer $l$ is generated by aggregating information from its 
ancestors $A = \{a_i^{l-1},a_i^{l-2},\ldots,a_i^{1},a_i^{0}\}$. However, we have 
empirically found that the information from up to three immediate ancestors 
$A = \{a_i^{l-1},a_i^{l-2},a_i^{l-3}\}$ is sufficient for realistic point cloud 
generation. This observation improves the overall computational cost of our 
network as $G_{PT}$ is not bounded by the entire depth of the tree as in 
\cite{shu20193d} (additional details and an analysis are included in Section 
\ref{sec:experiments}). Therefore, the output of $(l+1)$-layer of $G_{PT}$ is a 
first order approximation of the Chebyshev expansion 
\begin{equation*}
  p_i^{l+1} = \sigma\Bigg(\textbf{F}_m^l(p_i^l) + \sum_{q_j \in A(p_i^l)} W_j^lq_j^l + b^l\Bigg),
\end{equation*}
where
\begin{equation*}
  \textbf{F}_m^l(p_i^l) = \sum_{j=1}^{m}S_j p_i^l + r_j,
\end{equation*} 
$\sigma(\cdot)$ is an activation function, $p_i^l$ is the $i$-th node of the 
graph at the $l$-layer, and $q_j^l$ is the $j$-th ancestor of $p_i^l$ from the 
set of three immediate ancestors of $p_i^l$. $W, b, S, r$ are learnable 
parameters and $\textbf{F}_m$ is a sub-network with $m$ support.

\begin{figure}[t]
\begin{center}
  \includegraphics[width=0.9\linewidth]{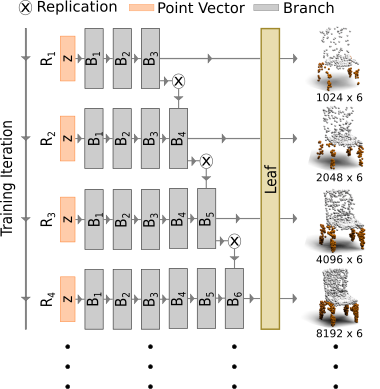}
\end{center}
\caption{The progressive growing of the point transformer $G_{PT}$. The network 
starts with a leaf output layer and branches $(B_1\ldots,B_T)$, each 
incorporating a tree graph, and converts a point vector $z$ into a 
low-resolution point cloud. After a predefined number of iterations, the number 
of branches is increased through the replication $R$ of the last branch $B_T$. 
In this figure, $T$ was kept small for visualization purposes. The process is 
continued until the generation of a point cloud at the desired resolution is 
achieved. Best viewed in color.}
\label{fig:pcgan_progressive_growing}
\end{figure}

Figure \ref{fig:pcgan_progressive_growing} presents an overview of the 
progressive growing of PCGAN. We have sub-divided the point transformer $G_{PT}$ 
into two parts, a branch $B$ and a leaf $L$, where a leaf acts as the output 
layer of $G_{PT}$ for increasing replications $R$. Each branch incorporates a 
tree graph of expanding depth, $H = (h_1,\ldots,h_T)$, where $T$ is the total 
number of branches. The leaf also incorporates a tree graph of depth $h_L$. 
First, we optimize the generator to produce a point cloud $\hat{x}$ at a 
predefined base resolution $(N_{R_1} \times 6)$ where 
$R_1 = h_L \Pi_{i=1}^T h_i$. After a fixed number of iterations, we introduce a 
new branch $B_{T+1}$ through the replication of the branch $B_{T}$ and we 
increase the depth $H = (h_1,\ldots,h_{T+1})$ of the incorporated tree graph. 
The replication of an already optimized branch facilitates the generation of 
higher resolution point clouds without starting the learning process from 
scratch. We continue this procedure until the desired resolution of the point 
cloud is realized, i.e., $R_{d} = h_L \Pi_{i=1}^{T+d} h_i$.

\subsection{Discriminator}
Given a point cloud $x$, the discriminator $D$ tries to predict the probability 
of $x$ being real or synthesized. The discriminator is comprised of the
following three sub-networks: a feature transformer $D_{FT}$, a label 
transformer $D_{LT}$, and a critic $D_{CR}$. The feature transformer network was 
inspired by \cite{wang2019dynamic} to collect local and global features through
a dynamic graph construction. Nonetheless, we expand the feature size in every
layer to account for the color of the point clouds. 

Given an input point vector, each layer of the point transformer network 
constructs a dynamic $k$-NN graph, 
$\mathcal{G}^l = (\mathcal{V}^l,\mathcal{E}^l)$, with self loops. Each point of 
the point cloud represents a vertex in the graph and the vertex of each 
subsequent layer depends on the output of the preceding layers. For a graph 
$\mathcal{G}^l$ at layer $l$, the edges $e_i^l$ between a vertex $v_i^l$ 
and its $k$ nearest neighbors are defined as a nonlinear function $\Theta = 
[\theta_1,\ldots,\theta_m]$ with learnable parameter $\theta$,
\begin{equation*}
  e_{i,(1,\ldots,k)}^l = \Theta_{j \in (1,\ldots,k)}(v_i^l,v_j^l).
\end{equation*}
Although there are three choices available for $\Theta$, we use the function 
defined in \cite{wang2019dynamic} to capture both local and global features. 
Concretely, for two vertices $v_i$ and $v_j$ we have
\begin{equation*}
  \Theta(v_i,v_j) = \theta(v_i,v_i-v_j). 
\end{equation*}
Lastly, the output of the feature transform network is a feature vector $f_c$ 
collected from the channel-wise maximum operation on the edge features from all
the edges of each vertex,
\begin{equation*}
  f_c = \max_{(i,j)\in\mathcal{E}}\Theta(v_i,v_j). 
\end{equation*}

The label transformer of the discriminator functions in an analogous way to the
label transformer of the generator and it provides a label vector $d_c$. Note 
that even though $D_{LT}$ and $G_{LT}$ are similar in architecture, their 
objectives are distinct. Hence, we use two different networks for this reason.
The feature vector $f_c$ together with the label vector $d_c$ are fed to the 
critic, a fully-connected three layer sub-network that aspires to predict the 
probability of its input vector being real or synthetic given that it was 
collected from object class $c$, i.e.,  
\begin{equation*}
  D_{CR}(f_c,d_c) = [P(\text{real}\,|\,c), P(\text{generated}\,|\,c)].
\end{equation*}

\vspace{-1mm}
\section{Experiments} 
\vspace{-1mm}
\label{sec:experiments}
In this section, we describe our experimental setup and provide an analysis of 
the results.

\textbf{Dataset.} We used the synthetic dataset ShapeNetCore 
\cite{chang2015shapenet} to conduct our experiments. ShapeNetCore is a 
collection of CAD models of various object classes among which we chose Chair, 
Table, Sofa, Airplane, and Motorcycle for our experiments. We have selected 
these classes for the diversity of their shapes and we keep the number of 
classes to five to reduce the training time. In addition, we have eliminated any 
CAD model that does not have material/color information. The training data was 
prepared by collecting point clouds of desired resolution from the surface of 
the CAD models. We normalized each point cloud such that the object is 
centralized in a unit cube and the RGB colors are in the range $[-0.5, +0.5]$. 
Table \ref{tab:training_data} shows the number of sample objects in each class.

\begin{table}
\begin{tabularx}{\columnwidth}{| X | X |} \hline
  Class Label & Number of Samples\\ \hline
  Airplane    & 4029\\ 
  Chair       & 4064\\ 
  Table       & 8474\\ 
  Sofa        & 3149\\ 
  Motorcycle  & 333\\ \hline
\end{tabularx}
\caption{The class labels and the number of samples in the training data.}
\label{tab:training_data}
\end{table}

\textbf{Implementation Details.} The generator of PCGAN learns to produce 
colored point clouds from low to high resolutions. We start the generation 
process with $N_{R_1} = 1024$ points and double the number of points 
progressively. To save memory and reduce computation time, we fix $N_{R_d} = 
8192$ points as the highest resolution for the point clouds. The input to the 
generator is a vector, $z_{in} \in \mathbb{R}^{64}$, sampled from a normal 
distribution $z_{in} \in \mathcal{N}(0,1)$, and the labels of the chosen 
classes $c \in C$. 

For the point transformer, the number of branches was set to $T = 5$ 
with depth increments $H = [1,2,2,2,2]$. The leaf increments were set to $h_L = 
64$. The depth increment hyperparameter for the branches and the leaves is 
similar to the branching of \cite{shu20193d}. The feature dimension for the 
layers of $G_{PT}$ was set to $[128,128,256,256,128,128,6]$. We used the Xavier 
initialization \cite{glorot2010understanding} for $G_{PT}$ and the support $q$ 
for each branch was set to $10$ as in \cite{shu20193d}.

The feature dimension for the layers of the feature transform sub-network were 
set to $[6,64,128,256,512,1024]$. To comply with the constraints of WGAN-GP, we 
did not use any batch normalization or dropout in $D_{FT}$. For $D_{FT}$, $k=20$ 
was used to construct the $k$-NN graph and $k$ was increased by $10$ with each 
increment in resolution.

For both the generator and the discriminator, LeakyReLU nonlinearity with a 
negative slope of 0.2 was employed. The learning rate $\alpha$ was set to 
$10^{-4}$ and the Adam optimizer \cite{kingma2014adam} with coefficients 
$\beta_1 = 0.0$ and $\beta_2 = 0.95$ were used. The gradient penalty coefficient 
$\lambda$ for WGAN-GP was set to 10.

\textbf{Metrics.} To quantitatively evaluate generated samples in 2D, the 
Fr\`echet inception distance \cite{heusel2017gans} is the most commonly used 
metric. We propose a similar metric called the Fr\`echet dynamic distance (FDD) 
to evaluate the generated point clouds where DGCNN \cite{wang2019dynamic} is 
used as a feature extractor. Although similar metrics exist \cite{shu20193d, 
sun2020pointgrow} where PointNet is used to extract features, the color of the 
point clouds is not considered and they suffer from the limitations of PointNet. 

We use DGCNN because it collects both local and global information over the 
feature space and it also performs better than PointNet in point cloud 
classification \cite{wang2019dynamic}. To implement FDD, we trained DGCNN until 
a 98\% test accuracy per class was achieved on the task of classification. 
Then, we extracted a $512$-dimensional feature vector from the average pooling 
layer of DGCNN to calculate the mean vector and covariance matrix. For real 
point clouds $x$ and synthetic point clouds $\hat{x}$, the FDD calculates the 
2-Wasserstein distance, 
\begin{equation*}
  \text{FDD}(x,\hat{x}) = \|\mu_x - \mu_{\hat{x}} \| + 
  \tr(\Sigma_x+\Sigma_{\hat{x}} - 2(\Sigma_x \Sigma_{\hat{x}})^{1/2}),
\end{equation*}
where $\mu$ and $\Sigma$ represent the mean vector and the covariance matrix,
respectively. The matrix trace is denoted by $\tr()$. Additionally, we have used 
the matrices from Achlioptas \etal \cite{achlioptas2018learning} for point cloud 
evaluation and we have compared the results with \cite{achlioptas2018learning,
shu20193d,valsesia2018learning}.

\begin{table*}
\begin{tabularx}{\textwidth}{| X | X | c c c c c|} \hline
Class      & Model & JSD $\downarrow$ & MMD-CD $\downarrow$ & MMD-EMD $\downarrow$ & COV-CD $\uparrow$ & COV-EMD $\uparrow$\\ \hline
           & r-GAN (dense) & 0.182 & 0.0009 & 0.094 & 31 & 9\\
           & r-GAN (conv) & 0.350 & \color{cyan}0.0008 & 0.101 & 26 & 7\\
Airplane   & Valsesia \etal (no up.) & 0.164 & 0.0010 & 0.102 & 24 & 13\\
           & Valsesia \etal (up.) & \color{magenta}{0.083} & \color{cyan}0.0008 & 0.071 & 31 & 14\\
           & tree-GAN & 0.097 & \color{magenta}{0.0004} & \color{magenta}0.068 & \color{magenta}61 & 20\\
           & {\bf PCGAN} (ours) & \color{cyan}0.085 & 0.0010 & \color{cyan} 0.070 & \color{cyan}37 & \color{magenta}29\\ \hline
    
           & r-GAN (dense) & 0.238 & 0.0029 &0.136 & \color{cyan}33 & 13\\
           & r-GAN (conv) & 0.517 & 0.0030 & 0.223 & 23 & 4\\
Chair      & Valsesia \etal(no up.) & 0.119 & 0.0033 & 0.104 & 26 & 20\\
           & Valsesia \etal (up.) & \color{cyan}0.100 & 0.0029 & \color{cyan}0.097 & 30 & 26\\
           & tree-GAN & 0.119 & \color{magenta}0.0016 & 0.101 & \color{magenta}58 & \color{cyan}30\\
           & {\bf PCGAN} (ours) & \color{magenta}0.089 & \color{cyan}0.0027 & \color{magenta}0.093 & 30 & \color{magenta}33\\ \hline
    
           & r-GAN (dense) & 0.221 & \color{magenta}0.0020 & 0.146 & \color{cyan}32 & 12\\
           & r-GAN(conv) & 0.293 & 0.0025 & 0.110 & 21 & 12 \\
Sofa       & Valsesia \etal (no up.) & \color{cyan}0.095 & \color{cyan}0.0024 & 0.094 & 25 & 19\\
           & Valsesia \etal (up.) & \color{magenta}0.063 & \color{magenta}0.0020 & \color{magenta}0.083 & \color{magenta}39 & \color{cyan}24\\
           & {\bf PCGAN} (ours) & 0.16 & 0.0027 & \color{cyan}0.093 & 24 & \color{magenta}27\\ \hline
    
Motorcycle & {\bf PCGAN} (ours) & 0.25 & 0.0016 & 0.097 & 10 & 9\\ \hline
  
Table      & {\bf PCGAN} (ours) & 0.093 & 0.0035 & 0.089 & 45 & 43\\ \hline
\end{tabularx}
\caption{A qualitative evaluation of the Jensen-Shannon divergence (JSD), 
the minimum matching distance (MMD), coverage (COV) with the Earth mover's 
distance (EMD), and the pseudo-chamfer distance (CD). Please refer to 
\cite{achlioptas2018learning} for details regarding the metrics. The results of 
previous studies are from \cite{shu20193d,valsesia2018learning}. The magenta and 
cyan values denote the best and the second best results, respectively. The 
resolution of the evaluated point clouds was $2048 \times 3$.}
\label{tab:metrics_results}
\end{table*}

\begin{table}
\begin{tabularx}{\columnwidth}{|X|c|c|} \hline
Class      & Real Data             & Generated Samples\\ \hline
Airplane   & $1.12 \times 10^{-5}$ & 4.58\\
Chair      & $2.07 \times 10^{-9}$ & 3.07\\ 
Motorcycle & $1.04 \times 10^{-4}$ & 13.25\\ 
Sofa       & $5.88 \times 10^{-6}$ & 3.14\\ 
Table      & $4.37 \times 10^{-7}$ & 2.02\\ \hline
\end{tabularx}
\caption{The FDD score for point cloud samples generated by PCGAN. Notice that 
the scores for real point clouds are almost zero. The point clouds were 
evaluated at a resolution of $8192 \times 6$.}
\label{tab:fdd_results}
\end{table}

\subsection{Results}
\begin{figure*}[p]
\begin{center}
  \includegraphics[width=0.90\linewidth]{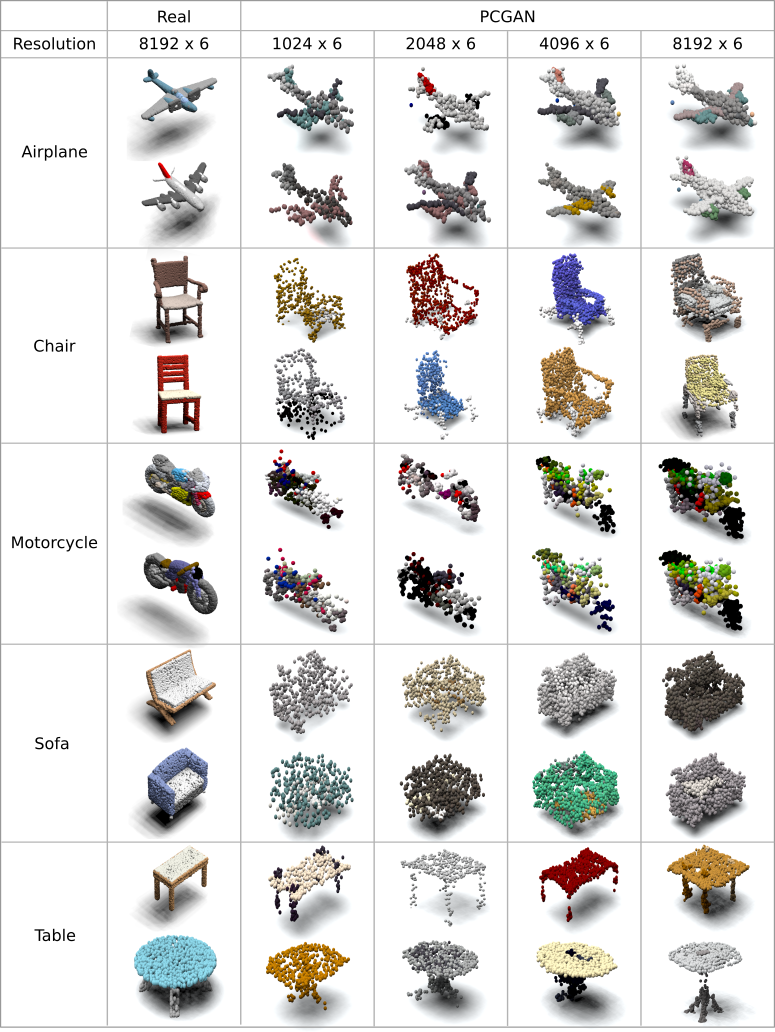}
\end{center}
\caption{The results of generated samples produced by PCGAN. Our model first 
learns the basic structure of an object at low resolutions and gradually builds 
up towards high-level details. The relationship between the object parts and
their colors (e.g., the legs of the chair/table are the same color while 
seat/top are contrasting) is also learned by the network. Mitsuba 2 
\cite{mitsuba2020} was used to render the point clouds. Best viewed in color.}
\label{fig:pcgan_results_2}
\end{figure*}

A set of synthesized objects along with their real counterparts is shown in 
Figure \ref{fig:pcgan_results_2}. As is evident from the samples, our model 
first learns the basic structure of an object in low resolutions and gradually 
builds up to higher resolutions. PCGAN also learns the relationship between 
object parts and color. For example, the legs of the chair have equal colors 
while the seat/arms are a different color, the legs of the table have matching 
colors while the top is a contrasting color, and the airplane wings/engine have 
the same color while the body/tail has a dissimilar color. 

\textbf{Quantitative Analysis.} We generated $5000$ random samples for each 
class and performed an evaluation using the matrices from
\cite{achlioptas2018learning}. Table \ref{tab:metrics_results} presents our 
findings along with comparisons to previous studies 
\cite{achlioptas2018learning,shu20193d,valsesia2018learning}. Note that 
although our model is capable of generating higher resolutions and colors, we 
only used point clouds with $N = 2048$ points in order to be comparable with 
other methods. Also, separate models were trained in 
\cite{achlioptas2018learning, shu20193d, valsesia2018learning} to generate 
point clouds of different classes while we have used the {\em same} model to 
generate point clouds for all five classes. Even though we have achieved 
comparable results in Table \ref{tab:metrics_results}, the main focus of our 
work is dense colored point cloud generation and point clouds with a resolution 
of $N = 2048$ is an intermediate result of our network. We have also evaluated 
colored dense point clouds ($N = 8192$) using the proposed FDD metric with the  
results presented in Table \ref{tab:fdd_results}.

\textbf{Computational Complexity.} Tree structured graph convolutions are used 
in each layer of the point transformer sub-network. Since every subsequent node 
in the tree is originally dependent upon all of its ancestors, the time 
complexity of the graph convolutions is 
$\sum_{l=1}^{L} B \times n_i^l \times A^l_i \times V^l_i$, where $B$ is the 
batch size of the training data, $L$ is the total number of layers, $n_i^l$ is
the height of the tree graph at the $i$-th node, $A_i$ is the induced number of 
ancestors preceding $i$-th node, and $V_i^l$ is the induced vertex number of the 
$i$-th node \cite{shu20193d}. However, we restrict $G_{PT}$ to aggregate 
information from at most three levels of ancestors in each layer. Thus, 
$n_i^l \le 3$ and the effective time complexity is 
$\sum_{l=1}^{L} B \times 3 \times A^l_i \times V^l_i$.

\textbf{Progressive Experiments.} We have experimented with different strategies 
of progressive growing such as branching $H = [2,4,4,4], [1,2,4,8], [1,2,4,2]$ 
and leaf $h_L = [4,16,64]$ increments. Although branching and leaf combinations
may generate more discernible point cloud geometries and colors for individual
classes, we have found that $H = [1,2,2,2]$ and $h_L =64$ work best in our 
experiments. Additionally, instead of creating new branches we have experimented 
with introducing new leaves for progressive growing. However, this seems to 
destabilize the network and results in the generation of inferior samples.

\textbf{Limitations and Future Work.} The main drawback of our model is the 
computational complexity. With $\approx 20000$ point clouds from five classes, 
our model takes roughly $15$ minutes per iteration (MPI) on four Nvidia GTX 1080
GPUs to generate point clouds $\hat{x} \in \mathbb{R}^{1024 \times 6}$. The MPI 
rises with every increase in resolution. For future work, we will attempt to 
reduce the computational complexity of our network. Our model also struggles
with generating objects that have fewer examples in the training data. The 
generated point clouds of the Motorcycle class in Figure 
\ref{fig:pcgan_results_2} is an example of this (ShapeNetCore has only $333$ CAD 
models of the Motorcycle). In the future, we will try to improve the 
generalization ability of our model and we will work on the addition of surface 
normal estimates.

\vspace{-1mm}
\section{Conclusion}
\vspace{-1mm}
\label{sec:conclusion}
This paper introduces PCGAN, the first conditional generative adversarial 
network to generate dense colored point clouds in an unsupervised mode. To 
reduce the complexity of the generation process, we train our network in a 
coarse to fine way with progressive growing and we condition our network on 
class labels for multiclass point cloud creation. We evaluate both point cloud 
geometry and color using our new FDD metric. In addition, we provide comparisons 
on point cloud geometry with recent generation techniques using available 
metrics. The evaluation results show that our model is capable of synthesizing 
high-quality point clouds for a disparate array of object classes. 

\ifthreedvfinal
\section*{Acknowledgments}
\vspace{-1mm}
The authors acknowledge the Texas Advanced Computing Center (TACC) at the
University of Texas at Austin for providing software, computational, and
storage resources that have contributed to the research results reported within
this paper.
\fi

{\small
\bibliographystyle{ieee}
\bibliography{a_progressive_conditional_generative_adversarial_network_for_generating_dense_and_colored_3d_point_clouds}

\begin{thebibliography}{10}\itemsep=-1pt

\bibitem{achlioptas2018learning}
P.~Achlioptas, O.~Diamanti, I.~Mitliagkas, and L.~Guibas.
\newblock Learning representations and generative models for 3d point clouds.
\newblock In {\em International Conference on Machine Learning}, pages 40--49,
  2018.

\bibitem{ahmed2018survey}
E.~Ahmed, A.~Saint, A.~E.~R. Shabayek, K.~Cherenkova, R.~Das, G.~Gusev,
  D.~Aouada, and B.~Ottersten.
\newblock A survey on deep learning advances on different 3d data
  representations.
\newblock {\em arXiv preprint arXiv:1808.01462}, 2018.

\bibitem{arjovsky2017wasserstein}
M.~Arjovsky, S.~Chintala, and L.~Bottou.
\newblock Wasserstein gan.
\newblock {\em arXiv preprint arXiv:1701.07875}, 2017.

\bibitem{atzmon2018point}
M.~Atzmon, H.~Maron, and Y.~Lipman.
\newblock Point convolutional neural networks by extension operators.
\newblock {\em arXiv preprint arXiv:1803.10091}, 2018.

\bibitem{biswas2012depth}
J.~Biswas and M.~Veloso.
\newblock Depth camera based indoor mobile robot localization and navigation.
\newblock In {\em IEEE International Conference on Robotics and Automation},
  pages 1697--1702, 2012.

\bibitem{bostelman2006applications}
R.~Bostelman, P.~Russo, J.~Albus, T.~Hong, and R.~Madhavan.
\newblock Applications of a 3d range camera towards healthcare mobility aids.
\newblock In {\em IEEE International Conference on Networking, Sensing and
  Control}, pages 416--421, 2006.

\bibitem{cao2019point}
X.~Cao and K.~Nagao.
\newblock Point cloud colorization based on densely annotated 3d shape dataset.
\newblock In {\em International Conference on Multimedia Modeling}, pages
  436--446. Springer, 2019.

\bibitem{cao2020psnet}
X.~Cao, W.~Wang, K.~Nagao, and R.~Nakamura.
\newblock Psnet: A style transfer network for point cloud stylization on
  geometry and color.
\newblock In {\em IEEE Winter Conference on Applications of Computer Vision},
  pages 3337--3345, 2020.

\bibitem{cao20173d}
Z.~Cao, Q.~Huang, and R.~Karthik.
\newblock 3d object classification via spherical projections.
\newblock In {\em International Conference on 3D Vision (3DV)}, pages 566--574.
  IEEE, 2017.

\bibitem{chang2015shapenet}
A.~X. Chang, T.~Funkhouser, L.~Guibas, P.~Hanrahan, Q.~Huang, Z.~Li,
  S.~Savarese, M.~Savva, S.~Song, H.~Su, et~al.
\newblock Shapenet: An information-rich 3d model repository.
\newblock {\em arXiv preprint arXiv:1512.03012}, 2015.

\bibitem{chaudhuri2019learning}
S.~Chaudhuri, D.~Ritchie, K.~Xu, and H.~Zhang.
\newblock Learning generative models of 3d structures.
\newblock {\em Eurographics Tutorial}, 2019.

\bibitem{eckart2016accelerated}
B.~Eckart, K.~Kim, A.~Troccoli, A.~Kelly, and J.~Kautz.
\newblock Accelerated generative models for 3d point cloud data.
\newblock In {\em Proceedings of the IEEE Conference on Computer Vision and
  Pattern Recognition}, pages 5497--5505, 2016.

\bibitem{gadelha2018multiresolution}
M.~Gadelha, R.~Wang, and S.~Maji.
\newblock Multiresolution tree networks for 3d point cloud processing.
\newblock In {\em Proceedings of the European Conference on Computer Vision
  (ECCV)}, pages 103--118, 2018.

\bibitem{glorot2010understanding}
X.~Glorot and Y.~Bengio.
\newblock Understanding the difficulty of training deep feedforward neural
  networks.
\newblock In {\em Proceedings of the Thirteenth International Conference on
  Artificial Intelligence and Statistics}, pages 249--256, 2010.

\bibitem{goodfellow2014generative}
I.~Goodfellow, J.~Pouget-Abadie, M.~Mirza, B.~Xu, D.~Warde-Farley, S.~Ozair,
  A.~Courville, and Y.~Bengio.
\newblock Generative adversarial nets.
\newblock In {\em Advances in Neural Information Processing Systems}, pages
  2672--2680, 2014.

\bibitem{gulrajani2017improved}
I.~Gulrajani, F.~Ahmed, M.~Arjovsky, V.~Dumoulin, and A.~C. Courville.
\newblock Improved training of wasserstein gans.
\newblock In {\em Advances in Neural Information Processing Systems}, pages
  5767--5777, 2017.

\bibitem{hertz2020pointgmm}
A.~Hertz, R.~Hanocka, R.~Giryes, and D.~Cohen-Or.
\newblock Pointgmm: a neural gmm network for point clouds.
\newblock In {\em Proceedings of the IEEE/CVF Conference on Computer Vision and
  Pattern Recognition}, pages 12054--12063, 2020.

\bibitem{heusel2017gans}
M.~Heusel, H.~Ramsauer, T.~Unterthiner, B.~Nessler, and S.~Hochreiter.
\newblock Gans trained by a two time-scale update rule converge to a local nash
  equilibrium.
\newblock In {\em Advances in Neural Information Processing Systems}, pages
  6626--6637, 2017.

\bibitem{hou20193d}
J.~Hou, A.~Dai, and M.~Nie{\ss}ner.
\newblock 3d-sis: 3d semantic instance segmentation of rgb-d scans.
\newblock In {\em Proceedings of the IEEE Conference on Computer Vision and
  Pattern Recognition}, pages 4421--4430, 2019.

\bibitem{ioannidou2017deep}
A.~Ioannidou, E.~Chatzilari, S.~Nikolopoulos, and I.~Kompatsiaris.
\newblock Deep learning advances in computer vision with 3d data: A survey.
\newblock {\em ACM Computing Surveys (CSUR)}, 50(2):1--38, 2017.

\bibitem{karras2017progressive}
T.~Karras, T.~Aila, S.~Laine, and J.~Lehtinen.
\newblock Progressive growing of gans for improved quality, stability, and
  variation.
\newblock {\em arXiv preprint arXiv:1710.10196}, 2017.

\bibitem{karras2019style}
T.~Karras, S.~Laine, and T.~Aila.
\newblock A style-based generator architecture for generative adversarial
  networks.
\newblock In {\em Proceedings of the IEEE Conference on Computer Vision and
  Pattern Recognition}, pages 4401--4410, 2019.

\bibitem{karras2020analyzing}
T.~Karras, S.~Laine, M.~Aittala, J.~Hellsten, J.~Lehtinen, and T.~Aila.
\newblock Analyzing and improving the image quality of stylegan.
\newblock In {\em Proceedings of the IEEE/CVF Conference on Computer Vision and
  Pattern Recognition}, pages 8110--8119, 2020.

\bibitem{kingma2014adam}
D.~P. Kingma and J.~Ba.
\newblock Adam: A method for stochastic optimization.
\newblock {\em arXiv preprint arXiv:1412.6980}, 2014.

\bibitem{li2018point}
C.-L. Li, M.~Zaheer, Y.~Zhang, B.~Poczos, and R.~Salakhutdinov.
\newblock Point cloud gan.
\newblock {\em arXiv preprint arXiv:1810.05795}, 2018.

\bibitem{li2018so}
J.~Li, B.~M. Chen, and G.~Hee~Lee.
\newblock So-net: Self-organizing network for point cloud analysis.
\newblock In {\em Proceedings of the IEEE Conference on Computer Vision and
  Pattern Recognition}, pages 9397--9406, 2018.

\bibitem{liu20183d}
B.~Liu, M.~Guo, E.~Chou, R.~Mehra, S.~Yeung, N.~L. Downing, F.~Salipur,
  J.~Jopling, B.~Campbell, K.~Deru, et~al.
\newblock 3d point cloud-based visual prediction of icu mobility care
  activities.
\newblock In {\em Machine Learning for Healthcare Conference}, pages 17--29,
  2018.

\bibitem{luo20163d}
R.~C. Luo, V.~W. Ee, and C.-K. Hsieh.
\newblock 3d point cloud based indoor mobile robot in 6-dof pose localization
  using fast scene recognition and alignment approach.
\newblock In {\em International Conference on Multisensor Fusion and
  Integration for Intelligent Systems (MFI)}, pages 470--475. IEEE, 2016.

\bibitem{maturana2015voxnet}
D.~Maturana and S.~Scherer.
\newblock Voxnet: A 3d convolutional neural network for real-time object
  recognition.
\newblock In {\em IEEE/RSJ International Conference on Intelligent Robots and
  Systems (IROS)}, pages 922--928, 2015.

\bibitem{mo2019structurenet}
K.~Mo, P.~Guerrero, L.~Yi, H.~Su, P.~Wonka, N.~Mitra, and L.~J. Guibas.
\newblock Structurenet: Hierarchical graph networks for 3d shape generation.
\newblock {\em arXiv preprint arXiv:1908.00575}, 2019.

\bibitem{pcgan2020}
\url{https://github.com/robotic-vision-lab/Progressive-Conditional-Generative-Adversarial-Network}.

\bibitem{pfrunder2017real}
A.~Pfrunder, P.~V. Borges, A.~R. Romero, G.~Catt, and A.~Elfes.
\newblock Real-time autonomous ground vehicle navigation in heterogeneous
  environments using a 3d lidar.
\newblock In {\em IEEE/RSJ International Conference on Intelligent Robots and
  Systems (IROS)}, pages 2601--2608, 2017.

\bibitem{pohlmann2016evaluation}
S.~T. P{\"o}hlmann, E.~F. Harkness, C.~J. Taylor, and S.~M. Astley.
\newblock Evaluation of kinect 3d sensor for healthcare imaging.
\newblock {\em Journal of Medical and Biological Engineering}, 36(6):857--870,
  2016.

\bibitem{qi2017pointnet}
C.~R. Qi, H.~Su, K.~Mo, and L.~J. Guibas.
\newblock Pointnet: Deep learning on point sets for 3d classification and
  segmentation.
\newblock In {\em Proceedings of the IEEE Conference on Computer Vision and
  Pattern Recognition}, pages 652--660, 2017.

\bibitem{qi2017pointnet++}
C.~R. Qi, L.~Yi, H.~Su, and L.~J. Guibas.
\newblock Pointnet++: Deep hierarchical feature learning on point sets in a
  metric space.
\newblock In {\em Advances in Neural Information Processing Systems}, pages
  5099--5108, 2017.

\bibitem{ramasinghe2019spectral}
S.~Ramasinghe, S.~Khan, N.~Barnes, and S.~Gould.
\newblock Spectral-gans for high-resolution 3d point-cloud generation.
\newblock {\em arXiv preprint arXiv:1912.01800}, 2019.

\bibitem{mitsuba2020}
{Realistic Graphics Lab, EPFL}.
\newblock Mitsuba 2 renderer, 2020.
\newblock http://www.mitsuba-renderer.org.

\bibitem{sarmad2019rl}
M.~Sarmad, H.~J. Lee, and Y.~M. Kim.
\newblock Rl-gan-net: A reinforcement learning agent controlled gan network for
  real-time point cloud shape completion.
\newblock In {\em Proceedings of the IEEE Conference on Computer Vision and
  Pattern Recognition}, pages 5898--5907, 2019.

\bibitem{sharma2019learning}
G.~Sharma, E.~Kalogerakis, and S.~Maji.
\newblock Learning point embeddings from shape repositories for few-shot
  segmentation.
\newblock In {\em International Conference on 3D Vision (3DV)}, pages 67--75.
  IEEE, 2019.

\bibitem{shu20193d}
D.~W. Shu, S.~W. Park, and J.~Kwon.
\newblock 3d point cloud generative adversarial network based on tree
  structured graph convolutions.
\newblock In {\em Proceedings of the IEEE International Conference on Computer
  Vision}, pages 3859--3868, 2019.

\bibitem{sun2020pointgrow}
Y.~Sun, Y.~Wang, Z.~Liu, J.~Siegel, and S.~Sarma.
\newblock Pointgrow: Autoregressively learned point cloud generation with
  self-attention.
\newblock In {\em IEEE Winter Conference on Applications of Computer Vision},
  pages 61--70, 2020.

\bibitem{tchapmi2019topnet}
L.~P. Tchapmi, V.~Kosaraju, H.~Rezatofighi, I.~Reid, and S.~Savarese.
\newblock Topnet: Structural point cloud decoder.
\newblock In {\em Proceedings of the IEEE Conference on Computer Vision and
  Pattern Recognition}, pages 383--392, 2019.

\bibitem{valsesia2018learning}
D.~Valsesia, G.~Fracastoro, and E.~Magli.
\newblock Learning localized generative models for 3d point clouds via graph
  convolution.
\newblock In {\em International Conference on Learning Representations}, 2018.

\bibitem{villani2008optimal}
C.~Villani.
\newblock {\em Optimal transport: old and new}, volume 338.
\newblock Springer Science \& Business Media, 2008.

\bibitem{wang2019dynamic}
Y.~Wang, Y.~Sun, Z.~Liu, S.~E. Sarma, M.~M. Bronstein, and J.~M. Solomon.
\newblock Dynamic graph cnn for learning on point clouds.
\newblock {\em ACM Transactions On Graphics ({TOG})}, 38(5):1--12, 2019.

\bibitem{wang2019forknet}
Y.~Wang, D.~J. Tan, N.~Navab, and F.~Tombari.
\newblock Forknet: Multi-branch volumetric semantic completion from a single
  depth image.
\newblock In {\em Proceedings of the IEEE International Conference on Computer
  Vision}, pages 8608--8617, 2019.

\bibitem{wang2018point}
Y.~Wang, S.~Zhang, B.~Wan, W.~He, and X.~Bai.
\newblock Point cloud and visual feature-based tracking method for an augmented
  reality-aided mechanical assembly system.
\newblock {\em The International Journal of Advanced Manufacturing Technology},
  99(9-12):2341--2352, 2018.

\bibitem{wen2020point}
X.~Wen, T.~Li, Z.~Han, and Y.-S. Liu.
\newblock Point cloud completion by skip-attention network with hierarchical
  folding.
\newblock In {\em Proceedings of the IEEE/CVF Conference on Computer Vision and
  Pattern Recognition}, pages 1939--1948, 2020.

\bibitem{whitty2010autonomous}
M.~Whitty, S.~Cossell, K.~S. Dang, J.~Guivant, and J.~Katupitiya.
\newblock Autonomous navigation using a real-time 3d point cloud.
\newblock In {\em Australasian Conference on Robotics and Automation}, pages
  1--3, 2010.

\bibitem{wu2018learning}
J.~Wu, C.~Zhang, X.~Zhang, Z.~Zhang, W.~T. Freeman, and J.~B. Tenenbaum.
\newblock Learning shape priors for single-view 3d completion and
  reconstruction.
\newblock In {\em Proceedings of the European Conference on Computer Vision
  (ECCV)}, pages 646--662, 2018.

\bibitem{wu2019pointconv}
W.~Wu, Z.~Qi, and L.~Fuxin.
\newblock Pointconv: Deep convolutional networks on 3d point clouds.
\newblock In {\em Proceedings of the IEEE Conference on Computer Vision and
  Pattern Recognition}, pages 9621--9630, 2019.

\bibitem{xie2020generative}
J.~Xie, Y.~Xu, Z.~Zheng, S.-C. Zhu, and Y.~Nian~Wu.
\newblock Generative pointnet: Energy-based learning on unordered point sets
  for 3d generation, reconstruction and classification.
\newblock {\em arXiv}, pages arXiv--2004, 2020.

\bibitem{yang2019pointflow}
G.~Yang, X.~Huang, Z.~Hao, M.-Y. Liu, S.~Belongie, and B.~Hariharan.
\newblock Pointflow: 3d point cloud generation with continuous normalizing
  flows.
\newblock In {\em Proceedings of the IEEE International Conference on Computer
  Vision}, pages 4541--4550, 2019.

\bibitem{yang2018foldingnet}
Y.~Yang, C.~Feng, Y.~Shen, and D.~Tian.
\newblock Foldingnet: Point cloud auto-encoder via deep grid deformation.
\newblock In {\em Proceedings of the IEEE Conference on Computer Vision and
  Pattern Recognition}, pages 206--215, 2018.

\bibitem{yi2017syncspeccnn}
L.~Yi, H.~Su, X.~Guo, and L.~J. Guibas.
\newblock Syncspeccnn: Synchronized spectral cnn for 3d shape segmentation.
\newblock In {\em Proceedings of the IEEE Conference on Computer Vision and
  Pattern Recognition}, pages 2282--2290, 2017.

\bibitem{yi2019gspn}
L.~Yi, W.~Zhao, H.~Wang, M.~Sung, and L.~J. Guibas.
\newblock Gspn: Generative shape proposal network for 3d instance segmentation
  in point cloud.
\newblock In {\em Proceedings of the IEEE conference on computer vision and
  pattern recognition}, pages 3947--3956, 2019.

\bibitem{yuan2018pcn}
W.~Yuan, T.~Khot, D.~Held, C.~Mertz, and M.~Hebert.
\newblock Pcn: Point completion network.
\newblock In {\em International Conference on 3D Vision (3DV)}, pages 728--737.
  IEEE, 2018.

\bibitem{zaheer2017deep}
M.~Zaheer, S.~Kottur, S.~Ravanbakhsh, B.~Poczos, R.~R. Salakhutdinov, and A.~J.
  Smola.
\newblock Deep sets.
\newblock In {\em Advances in Neural Information Processing Systems}, pages
  3391--3401, 2017.

\bibitem{zhou2018voxelnet}
Y.~Zhou and O.~Tuzel.
\newblock Voxelnet: End-to-end learning for point cloud based 3d object
  detection.
\newblock In {\em Proceedings of the IEEE Conference on Computer Vision and
  Pattern Recognition}, pages 4490--4499, 2018.

\bibitem{zhu20123d}
Q.~Zhu, L.~Chen, Q.~Li, M.~Li, A.~N{\"u}chter, and J.~Wang.
\newblock 3d lidar point cloud based intersection recognition for autonomous
  driving.
\newblock In {\em IEEE Intelligent Vehicles Symposium}, pages 456--461, 2012.

\end{thebibliography}
}

\end{document}